\documentclass[11pt]{article}
\usepackage[a4paper]{geometry}
\usepackage{amsthm}
\usepackage{amsmath}
\usepackage{amssymb}

\usepackage{amsfonts}
\usepackage{graphicx}
\usepackage{color}
\usepackage[bf,SL,BF]{subfigure}
\usepackage{url}
\usepackage{epstopdf}
\usepackage{pdfsync}
\usepackage[colorlinks]{hyperref}
\usepackage[all]{xypic}
\usepackage{comment}
\usepackage{mathabx}
\usepackage{upgreek}

\hypersetup{
    linkcolor=blue,
}

\newlength{\fixboxwidth}
\usepackage{epsfig,amsbsy,graphicx,multirow}

\usepackage{ algorithm, algorithmic}
\renewcommand{\algorithmiccomment}[1]{\bgroup\hfill//~#1\egroup}

\usepackage[all]{xy}
\usepackage{accents}

 \numberwithin{equation}{section}

\def\R{\mathbb{R}}

\def\X{{\bf\mathcal{X}}}

\def\L{\mathcal{L}}

\def\restrict#1{\raise-.5ex\hbox{\ensuremath|}_{#1}}

\def\<{\big\langle}
\def\>{\big\rangle}

\definecolor{red}{rgb}{0.9, 0, 0}

\begin{document}
\title{ Deep regularization and direct training of the inner layers
of  Neural Networks with Kernel Flows}

\date{\today}

\author{Gene Ryan Yoo\thanks{Caltech, MC 253-37, Pasadena, CA 91125, USA, gyoo@caltech.edu} \and Houman Owhadi\thanks{Caltech,  MC 9-94, Pasadena, CA 91125, USA, owhadi@caltech.edu}}

\maketitle

\begin{abstract}
We introduce a new regularization method for Artificial Neural Networks (ANNs) based on Kernel Flows (KFs). KFs were introduced in \cite{owhadi2019kernel} as a method for kernel selection in regression/kriging based on the minimization of the loss of accuracy incurred by halving the number of interpolation points in random batches of the dataset.
 Writing   $f_\theta(x) = \big(f^{(n)}_{\theta_n}\circ f^{(n-1)}_{\theta_{n-1}} \circ \dots \circ f^{(1)}_{\theta_1}\big)(x)$ for the functional representation of compositional structure of the ANN (where $\theta_i$ are the weights and biases of the layer $i$), the inner layers outputs $h^{(i)}(x) = \big(f^{(i)}_{\theta_i}\circ f^{(i-1)}_{\theta_{i-1}} \circ \dots \circ f^{(1)}_{\theta_1}\big)(x)$ define a hierarchy of feature maps and a hierarchy of kernels $k^{(i)}(x,x')=\exp(- \gamma_i \|h^{(i)}(x)-h^{(i)}(x')\|_2^2)$.
When combined with a batch of the dataset these kernels produce KF losses $e_2^{(i)}$ (defined as the $L^2$ regression error incurred by
using a random half of the batch to predict the other half) depending on the parameters of the inner layers $\theta_1,\ldots,\theta_i$ (and $\gamma_i$).
The proposed method simply consists in aggregating (as a weighted sum) a subset of these KF losses with a classical output loss (e.g. cross-entropy).
We test the proposed method on Convolutional Neural Networks (CNNs) and Wide Residual Networks (WRNs) without alteration of their structure nor their output classifier and  report reduced test errors, decreased generalization gaps, and increased robustness to distribution shift without significant increase in computational complexity relative to standard  CNN and WRN training (with Drop Out and Batch Normalization).
We suspect that these results might be explained by the fact that while conventional training only employs a linear functional (a generalized moment) of the empirical distribution defined by the dataset and can be prone to trapping in the Neural Tangent Kernel regime (under over-parameterizations), the proposed loss function  (defined as a nonlinear functional of the empirical distribution) effectively trains the underlying kernel defined by the CNN beyond regressing the data with that kernel.

\end{abstract}

\section{A reminder on Kernel Flows}

 Kernel Flows were introduced in \cite{owhadi2019kernel} as a method for kernel selection/design in Kriging/Gaussian Process Regression (GPR).
 As a reminder on KFs consider the problem of approximating an unknown function $u^\dagger$ mapping $\X$ to $\R$   based on the input/output dataset $(x_i,y_i)_{1\leq i \leq N}$ ($u^\dagger(x_i)=y_i$).
 Any non-degenerate kernel $K(x,x')$ can be used to approximate $u^\dagger$ with the interpolant
 \begin{equation}
 u(x)=K(x,X) K(X,X)^{-1} Y\,,
  \end{equation}
writing $Y:=(y_1,\ldots,y_N)^T$, $X:=(x_1,\ldots,x_N)$, $K(X,X)$ for the $N\times N$ Gram matrix $K(x_i,x_i)$ and $K(x,X)$ for the $N$ dimensional vector with entries $K(x,x_i)$.
 The kernel selection problem concerns the identification of a good kernel for performing this interpolation. The KF approach to this problem is to simply use the loss of accuracy incurred by removing half of the dataset as a loss of kernel selection. The application of this process to minibatches results in a loss that is doubly randomized by (1) the selection of the minibatch (2) the half sub-sampling of the minibatch. An iterated steepest descent minimization of this loss then results in stochastic  gradient descent algorithm (where the minibatch and its half-subset  are re-sampled at each step).
 Given a family of kernels $K_\theta(x,x')$ parameterized by $\theta$, the resulting algorithm can then be described as follows:
 (1) Select random subvectors $X^b$ and $Y^b$ of $X$ and $Y$ (through uniform sampling without replacement in the index set $\{1,\ldots,N\}$) (2) Select random subvectors $X^c$ and $Y^c$ of $X^b$ and $Y^b$ (by selecting, at random, uniformly and without replacement, half of the indices defining $X^b$) (3) Let $\rho(\theta,X^b,Y^b,X^c,Y^c)$ be the squared relative error (in the RKHS norm $\|\cdot\|_{K_\theta}$ defined by $K_\theta$)  between
 the interpolants $u^b$ and $u^c$ obtained from the two nested subsets of the dataset and the kernel $K_\theta$, i.e.\footnote{ $\rho:=\|u^b-u^c\|^2_{K_\theta}/\|u^b\|^2_{K_\theta}$, with $u^b(x)=K_\theta(x,X^b) K_\theta(X^b,X^b)^{-1} Y^b$ and $u^c(x)=K_\theta(x,X^c) K_\theta(X^c,X^c)^{-1} Y^c$, and $\rho$  admits \cite[Prop.~13.29]{owhadi2019operator} the representation \eqref{eqjehdhebdhdhj} enabling its computation}
 \begin{equation}\label{eqjehdhebdhdhj}
 \rho(\theta,X^b,Y^b,X^c,Y^c):=1-\frac{Y^{c,T} K_\theta(X^c,X^c)^{-1} Y_c}{Y^{f,T} K_\theta(X^b,X^b)^{-1} Y^b}\,.
 \end{equation}
 (4) evolve $\theta$ in the gradient descent direction of $\rho$, i.e. $\theta \leftarrow \theta - \delta \nabla_\theta \rho$ (5) repeat.

\begin{figure}[h]
\begin{center}
\includegraphics[width= 0.85 \textwidth]{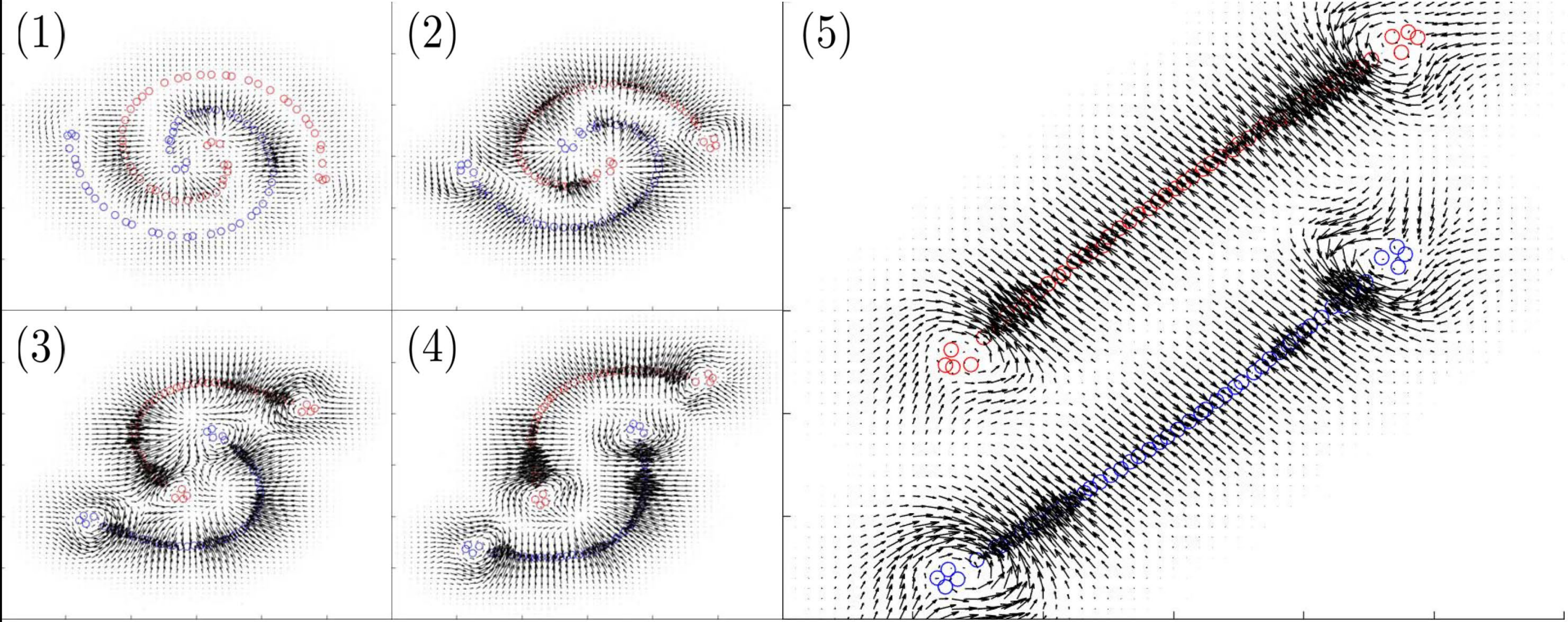}
\caption{\cite[Fig.~13]{owhadi2019kernel}, $(F_n(x_i))_{1\leq i \leq N}$ (dots) and $10(F_{n+300}(x)-F_n(x))/300$ (arrows) for 5 different values of $n$.
}
\label{figswissroll7}
\end{center}
\end{figure}

\paragraph{Example.}
Fig.~\ref{figswissroll7} shows an application of the proposed approach to the selection of a kernel
$K_F(x,x')=\exp(-\gamma \|F(x)-F(x')\|^2 )$ parameterised by a deformation $F\,:\,\R^2\rightarrow \R^2$ of the input space ($\X=\R^2$).
The dataset is the swissroll cheesecake (red points have labels $+1$ and blue points have labels $-1$), Fig.~\ref{figswissroll7} shows the deformed dataset $F_n(X)$ and the gradient $-\nabla_F \rho$  averaged over $300$ steps.

\paragraph{The $l_2$-norm variant.}
In this paper we will consider the $l_2$-norm variant of KF (introduced in \cite[Sec.~10]{owhadi2019kernel}) in which the instantaneous loss $\rho$ in \eqref{eqjehdhebdhdhj} is replaced by the error (let $\|\cdot\|_2$ be the Euclidean $l_2$  norm) $e_2:=\|Y^b - u^c(X^b)\|_2^2$ of $u^c$ in predicting the labels $Y^b$, i.e.
\begin{equation}
e_2(\theta,X^b,Y^b,X^c,Y^c):=\|Y^b - K_\theta(X^b,X^c) K_\theta(X^c,X^c)^{-1} Y^c\|_2^2
\end{equation}

\section{Kernel Flow regularization of Neural Networks}\label{KF reg}
Write
\begin{equation}
    f_\theta(x) = \big(f^{(n)}_{\theta_n}\circ f^{(n-1)}_{\theta_{n-1}} \circ \dots \circ f^{(1)}_{\theta_1}\big)(x)
\end{equation}
for the compositional structure of an artificial neural network (ANN) with input $x$ and $n$ layers
$ f^{(i)}_{\theta_i}(z) = \phi(W_i z + b_i)$ parameterized by the weights and biases $\theta_i:=(W_i,b_i)$, $\theta := \{\theta_1, \dots, \theta_n\}$. We will use ReLU for the  non-linearity $\phi$ in our experiments.
For  $i\in\{1, \dots, n-1\}$ let $ h^{(i)}(x)$ be the output of the $i$-th (inner)  layer, i.e.
\begin{equation}
    h_\theta^{(i)}(x) := \big(f^{(i)}_{\theta_i}\circ f^{(i-1)}_{\theta_{i-1}} \circ \dots \circ f^{(1)}_{\theta_1}\big)(x) \, ,
\end{equation}
 and let  $h_\theta(x) := (h_\theta^{(1)}(x), \dots, h_\theta^{(n-1)}(x))$ be the $(n-1)$-ordered tuple representing all inner layer outputs.
Let $k_{\gamma}(\cdot, \cdot)$  be a family of kernels parameterized by $\gamma$ and let $K_{\gamma,\theta}$ be the family of kernels parameterized by $\gamma$ and $\theta$ defined by
\begin{equation}\label{eqgjyguyuyb}
    K_{\gamma,\theta}(x, x') = k_{\gamma}(h_\theta(x), h_\theta(x'))\, .
\end{equation}

Given the random mini-batch $(X^b,Y^b)$ let  $\L_\text{c-e}(f_\theta(X^b), Y^b):=\sum_i \L_\text{c-e}(f_\theta(X^b_i), Y^b_i)$ be the cross-entropy loss associated with that mini-batch. Given the (randomly sub-sampled) half sub-batch $(X^c,Y^c)$,
 let $ \L_\text{KF}(\gamma,\theta,X^b,Y^b, X^c,Y^c)$ be the loss function (with   hyper-parameter $\lambda\geq 0$) defined by
\begin{equation}\label{KFloss}
    \L_\text{KF}: = \lambda \|Y^b - K_{\gamma,\theta}(X^b,X^c) K_{\gamma,\theta}(X^c,X^c)^{-1} Y^c\|_2^2
     +  \L_\text{c-e}(f_\theta(X^b), Y^b)\,.
\end{equation}

Our proposed KF-regularization approach is then to train the parameters $\theta$ of the network $f_\theta$ via the steepest descent $(\gamma,\theta)\leftarrow (\gamma,\theta) - \delta \nabla_{\gamma,\theta} \L_\text{KF}$.
Note that this algorithm (1) is randomized through both the sampling of the minibatch and its subsampling (2) adapts both $\theta$ and $\gamma$ (since the KF term depends on both $\theta$ and $\gamma$)  (3) simultaneously trains the accuracy of the output via the cross-entropy term and the generalization properties of the feature maps defined by the inner layers via the KF term. Furthermore while  the cross-entropy term is a linear functional of the empirical distribution $\frac{1}{N_b}\sum_i \updelta_{(X^b_i,Y^b_i)} $ defined by the mini-batch (writing $N_b$ for the number of indices contained in the mini-batch), the KF term is non-linear.
While $K_{\gamma,\theta}$ may depend on the output of all the inner layers, in our numerical experiments we have restricted its
dependence to the output of only one inner layer or used a weighted sum of such terms.

\section{Numerical experiments}

We will now use the  proposed KF regularization method to train a simple Convolutional Neural Network (CNN) on MNIST and Wide Residual Networks (WRN) \cite{WRNZagKom} on fashion MNIST, CIFAR-10, and CIFAR-100. Our goal is to test the proposed approach and compare its performance with popular ones (Batch Normalization and Drop Out).

\subsection{Kernel Flow regularization on MNIST}

We consider a Convolutional Neural Network (CNN) with six convolutional layers and three fully connected layers, as charted in Table \ref{MNIST arch} (this CNN is a variant of a CNN presented in \cite{Kagglemnist} with code used from \cite{CNNTF}).  Convolutional layers all have stride one in this network with the number of convolutional channels and the convolutional kernel size in the second and third columns from the left.  ``Valid'' padding implies no 0-padding at the boundaries of the image while ``same'' 0-pads images to obtain convolutional outputs with the same sizes as the inputs.  The ``Max Pool'' layers down sample their inputs by reducing each $2 \times 2$ square to their maximum values.  The ``Average Pool'' layer in the final convolutional layer takes a simple mean over each channel.  The final three layers are fully connected each with outputs listed on the right column.  All convolutional and dense layers include trainable biases.  Using notations from the previous section, the outputs of the convolutional layers, which include ReLU and pooling, are  $h^{(1)}(x)$ to $h^{(6)}(x)$ with output shapes  described in the left column.  The dense layers outputs are  $h^{(7)}(x)$ to $h^{(9)}(x)$.  We do not pre-process the data and, when employed, the data augmentation step, in this context, passes the original MNIST image to the network with probability $\frac{1}{3}$, applies an elastic deformation \cite{elasticdef} with probability $\frac{1}{3}$, and a random small translation, rotation, and shear with probability $\frac{1}{3}$.  The learning rate begins at $10^{-2}$ and smoothly exponentially decreases to $10^{-6}$ while training over $20$ epochs.

\begin{table}
{\scriptsize{
\begin{center}
\begin{tabular}{ | p{3.8cm} || p{1.9cm} p{1.9cm} p{1.9cm} | p{3cm} |}
\hline
\textbf{Layer Type} & \textbf{Number of filters} & \textbf{Filter size} & \textbf{Padding} & \textbf{Output shape} \\
\hline
Input layer & & & & $28 \times 28 \times 1$\\
\hline
Convolutional layer 1, ReLU & $150$ & $3 \times 3$ & Valid & $26 \times 26 \times 150$\\
\hline
Convolutional layer 2, ReLU & $150$ & $3 \times 3$ & Valid & $24 \times 24 \times 150$\\
\hline
Convolutional layer 3, ReLU & $150$ & $5 \times 5$ & Same & $24 \times 24 \times 150$\\
Max Pool & & $2 \times 2$ & & $12 \times 12 \times 150$\\
\hline
Convolutional layer 4, ReLU & $300$ & $3 \times 3$ & Valid & $10 \times 10 \times 300$\\
\hline
Convolutional layer 5, ReLU & $300$ & $3 \times 3$ & Valid & $8 \times 8 \times 300$\\
\hline
Convolutional layer 6, ReLU & $300$ & $5 \times 5$ & Same & $8 \times 8 \times 300$\\
Max Pool & & $2 \times 2$ & & $4 \times 4 \times 300$\\
Average Pool & & $4 \times 4$ & & $300$\\
\hline
Dense layer 1, ReLU & & & & $1200$\\
\hline
Dense layer 2, ReLU & & & & $300$\\
\hline
Dense layer 3 & & & & $10$\\
\hline
Softmax Output layer & & & & $10$\\
\hline
\end{tabular}
\end{center}
}}
\caption{The architecture of the CNN used in KF regularization experiments is charted.  Convolutional layers are divided with horizontal lines.  The middle block shows layer specifics and the shapes of the outputs of each layer is on the right.}\label{MNIST arch}
\end{table}

\subsubsection{Comparisons to Dropout}
The first experiment we present is one comparing the use of our KF loss function with the use of dropout (DO) \cite{dropout}.  We use Batch Normalization (BN) \cite{batchnorm} in all the experiments, i.e. both DO and KF regularization, in this subsection.  Our first dropout experiment uses dropping probabilities of $0.25$ across all layers. Our second experiment uses  dropping probabilities of $0.4$ over convolutional layers and $0.2$ on the dense layers.  We denote these as DO $0.25/0.25$ and DO $0.4/0.2$ respectively.

\begin{table}[h]
\begin{center}
\begin{tabular}{ | p{3.2cm} || p{3.2cm} | p{3.2cm} | p{3.2cm}|}
\hline
\textbf{Training Method} & \textbf{Original MNIST} & \textbf{Data augmented} & \textbf{QMNIST}\\
\hline
BN only & $0.500 \pm 0.054\%$ & $0.357 \pm 0.029\%$ & $0.453 \pm 0.025\%$\\
\hline
BN+DO $0.25/0.25$ & $0.403 \pm 0.048\%$ & $0.315 \pm 0.033\%$ & $0.429 \pm 0.012\%$ \\
\hline
BN+DO $0.4/0.2$ & $0.395 \pm 0.036\%$ & $0.331 \pm 0.031\%$ & $0.443 \pm 0.018\%$\\
\hline
BN+KF $6$ & $0.305 \pm 0.028\%$ & $0.281 \pm 0.026\%$ & $0.343 \pm 0.012\%$ \\
\hline
BN+KF $3,6$ & $0.307 \pm 0.027\%$ & $0.276 \pm 0.027\%$ & $0.356 \pm 0.012\%$ \\
\hline
\end{tabular}
\end{center}
\caption{A comparison of the average and standard deviation of testing errors each over 20 runs for networks.  The first data column on the left shows networks trained and tested on original MNIST data.  The middle is trained using data augmentation and uses original MNIST testing data.  The right column shows the same data augmented trained network, but uses QMNIST testing data \cite{qmnist}. }\label{KFvsDO}
\end{table}

We present two KF experiments.  The first one involves the following Gaussian kernel on the final convolutional layer $h^{(6)}(x) \in \R^{300}$:
\begin{equation}
    K^{(6)}_{\gamma_6,\theta}(x, x') = k^{(6)}_{\gamma_6} (h^{(6)}(x), h^{(6)}(x')) = e^{-\gamma_6 \| h^{(6)}(x) - h^{(6)}(x') \|^2}\, .
\end{equation}
We optimize the loss function in \eqref{KFloss} with kernel $K^{(6)}_{\gamma_6}$ over the parameters $\theta$ and $\gamma_6$.  The second experiment is a slight variant where we use both $K^{(6)}$ and
\begin{equation}
    K^{(3)}_{\gamma_3,\theta}(x, x') = k^{(3)}_{\gamma_3} (h^{(3)}(x), h^{(3)}(x')) = e^{-\gamma_3 \| a(h^{(3)}(x)) - a(h^{(3)}(x')) \|^2}\, ,
\end{equation}
where $a$ is a $12 \times 12$ average pooling reducing each channel to a single point.

Given the random mini-batch $(X^b,Y^b)$ and the (randomly sub-sampled) half sub-batch $(X^c,Y^c)$, we evolve $\theta$ and $\gamma_6$ in the steepest descent direction of the loss
{\small
\begin{equation}\label{KFloss2}
\begin{split}
    \L_\text{KF3,6} =& \lambda^{(3)} \|Y^b - K^{(3)}_{\gamma_3,\theta}(X^b,X^c) K^{(3)}_{\gamma_3,\theta}(X^c,X^c)^{-1} Y^c\|_2^2
     \\&+ \lambda^{(6)} \|Y^b - K^{(6)}_{\gamma_6,\theta}(X^b,X^c) K^{(6)}_{\gamma_6,\theta}(X^c,X^c)^{-1} Y^c\|_2^2 + \L_\text{c-e}(f_\theta(X^b), Y^b)
\end{split}
\end{equation}}
with respect to $\theta$, $\gamma_6$, and $\gamma_3$.  These two training methods are labeled KF $6$ and KF $3,6$ respectively.  The comparison between dropout and KF is made in table \ref{KFvsDO}.  KF $6$ and the network architecture was inspired by the work in \cite[Sec. 10]{owhadi2019kernel} (the GPR estimator on the final convolutional output space is here replaced by a fully connected network to minimize computational complexity).  On a $12$GB NVIDIA GeForce GTX TITAN X graphics card, training one dropout network ($20$ epochs) takes $1605$s to run, compared with $1629$s for KF $6$ and $1638$s for KF $3,6$.   Furthermore, this KF framework has another advantage of being flexible, allowing the control of generalization properties of multiple layers of the network simultaneously, as in KF $3, 6$, which does so on the outputs of the max pooling convolutional layers.

For each of the training methods we experiment with using original MNIST training and testing data, augmenting the MNIST training set and testing on the original data, and finally training on the augmented set, but testing on QMNIST, which is resampled MNIST test data \cite{qmnist}.  These three regimes are presented in the data columns of table \ref{KFvsDO} from left to right.  The difference between the original data augmented and QMNIST training errors quantifies the effect of distributional shift of the training data \cite{indistshift}.  This effect is observed to be reduced when using KF trained networks, which suggests some degree of robustness to distributional shift.
\begin{figure}[h!]
	\begin{center}
			\includegraphics[width=\textwidth]{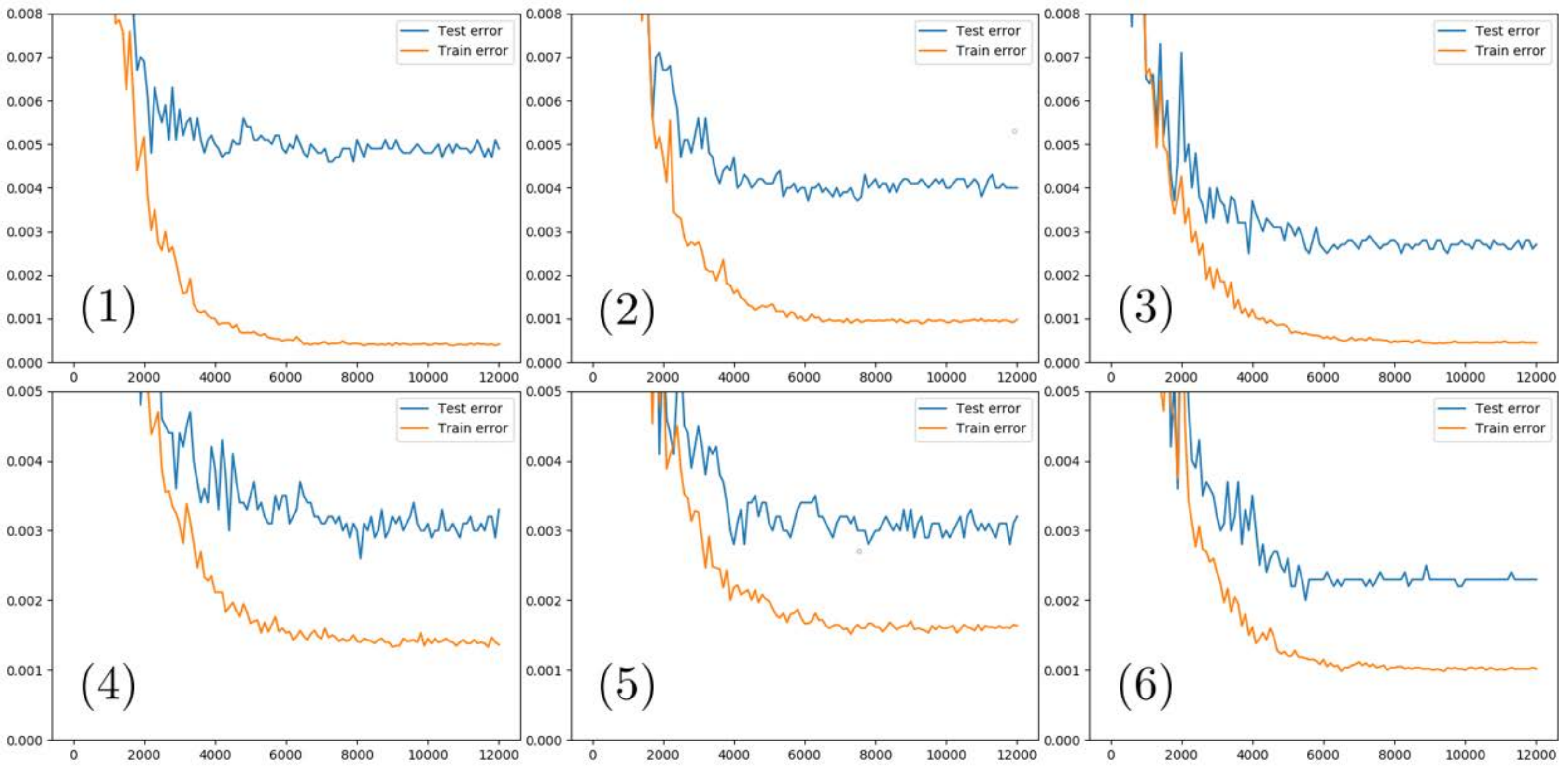}
		\caption{Training and testing errors are plotted over single runs trained with original data using (1) batch normalization only (2) dropout $0.25/0.25$ (3) KF $3, 6$.  Data augmented trained network errors are shown using (4) batch normalization only (5) dropout $0.25/0.25$ (6) KF $3,6$.}\label{gengap}
	\end{center}
\end{figure}
The training and testing errors of single runs with the batch normalization only, dropout $0.25/0.25$ and KF $3,6$ are plotted in figure \ref{gengap}.  Observe that the generalization gap (the gap between the training and testing errors) decreases with the use of dropout and that decrease is even more pronounced with KF.  Furthermore, contrary to dropout,  KF does not reduce training accuracy.  We observe similar findings on networks trained using data augmentation, albeit less pronounced.
\begin{figure}[h!]
	\begin{center}
			\includegraphics[width=\textwidth]{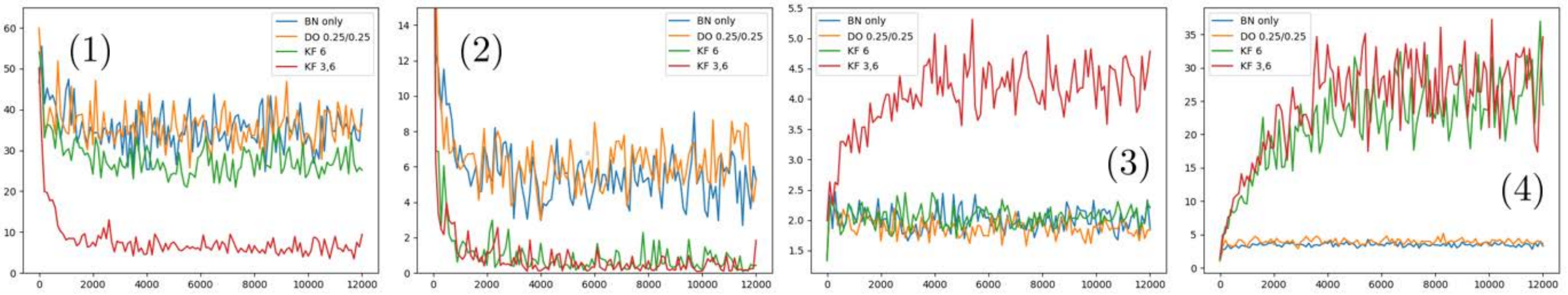}
		\caption{Single run over each of BN only, DO $0.25/0.25$, KF $6$, and KF$3,6$ training methods plotting (1) 3rd layer KF-loss (2) 6th layer KF-loss (3) ratio of mean inter-class and in-class distances of 3rd layer outputs (4) ratio of mean inter-class and in-class distances of 6th layer outputs.}\label{KF_lossratio}
	\end{center}
\end{figure}
We finally examine the components of the KF $3,6$ loss function as in equation \eqref{KFloss2}.  The KF-loss at the 3rd layer, $ \|Y^b - K^{(3)}_{\gamma_3,\theta}(X^b,X^c) K^{(3)}_{\gamma_3,\theta}(X^c,X^c)^{-1} Y^c\|_2^2$, and the 6th layer, $\|Y^b - K^{(6)}_{\gamma_6,\theta}(X^b,X^c) K^{(6)}_{\gamma_6,\theta}(X^c,X^c)^{-1} Y^c\|_2^2$, is computed for batch normalization, dropout, and KF training in figure \ref{KF_lossratio}.  It can be seen that KF $6$ reduces the 3rd layer KF-loss slightly compared to BN or DO $0.25/0.25$, while significantly reducing the 6th layer KF-loss.  Additionally, as expected, KF $3,6$ reduces both.  We can further consider the ratio of mean inter-class and in-class distances within each batch of 3rd and 6th convolutional layer outputs.  We see that KF $6$ separates images based on class in the 6th layer outputs while KF $3,6$ does so on both the 3rd and 6th.

\subsection{Kernel Flow regularization on Fashion MNIST and CIFAR}

\begin{table}
{\scriptsize
\begin{center}
\begin{tabular}{ | p{3.8cm} || p{1.9cm} p{1.9cm} p{2.3cm} | p{2.5cm} |}
\hline
\textbf{Layer/Block name} & \textbf{Number of filters} & \textbf{Filter size} & \textbf{Number of residual layers} & \textbf{Output shape} \\
\hline
Input layer & & & & $32 \times 32 \times 3$\\
\hline
Convolutional block 1 & $16$ & $3 \times 3$ & $1$ & $32 \times 32 \times 16$\\
\hline
Convolutional block 2 & $16k$ & $3 \times 3$ & $N$ & $32 \times 32 \times 16k$\\
\hline
Convolutional block 3 & $32k$ & $3 \times 3$ & $N$ & \\
Max Pool & & $2 \times 2$ & & $16 \times 16 \times 32k$\\
\hline
Convolutional block 4 & $64k$ & $3 \times 3$ & $N$ & \\
Max Pool & & $2 \times 2$ & & $8 \times 8 \times 64k$ \\
Average Pool & & $8 \times 8$ & & $64k$\\
\hline
Dense layer & & & & $64k$\\
\hline
Softmax Output layer & & & & $10$\\
\hline
\end{tabular}
\end{center}}
\caption{The architecture of the WRN used in KF regularization experiments with CIFAR input images.  Convolutional blocks are divided with horizontal lines.  The middle portion shows block specifics such as filter width and depth in each block and the shapes of the outputs of each layer is on the right.  Note that max pooling occurs within the last residual layer of each block.}\label{WRN arch}
\end{table}

We now consider the Wide Residual Network (WRN) structure  described in \cite[Table 1]{WRNZagKom} with the addition of a dense layer.  For convenience, we show this architecture in Table \ref{WRN arch}.  Note that there are four convolutional blocks, each with a certain number of residual layers, which are as described in \cite[Fig. 1c,d]{WRNZagKom} for BN and BN+DO training respectively.  Each layer consists of two convolutional blocks, with dropout applied between the blocks in dropout training, added to an identity mapping from the input of the layer.  In our dropout experiments, we drop each neuron in the network with probability $0.3$.  Note that $k$ and $N$ are hyper-parameters of the WRN architecture governing width and depth respectively, and a network with such $k, N$ is written WRN-$k$-$N$.  In these presented WRN experiments, we use data augmentation where training images are randomly translated and horizontally flipped.  In our implementations, we have modified the code from \cite{WRNTF} (which uses TensorFlow).  Batches consisting of $100$ images are used in these experiments.  In fashion MNIST and CIFAR-10, each half batch contains $5$ random images from each of the $10$ classes.  Meanwhile in CIFAR-100, we require each class represented in the testing sub-batch to also be represented in the training sub-batch. 

We write the outputs of each of the four convolutional blocks as $h^{(1)}(x), \dots, h^{(4)}(x)$.  Again defining $a$ as the average pooling operator, we have $a(h^{(1)}(x)) \in \R^{16}$, $a(h^{(2)}(x))\in \R^{16k}$, $a(h^{(3)}(x))\in\R^{32k}$, and $a(h^{(4)}(x)) = h^{(4)}(x)\in\R^{64k}$.  We define corresponding RBF kernels
\begin{equation}
    K^{(l)}_{\gamma_l}(x, x') = k^{(l)}_{\gamma_l} (h^{(l)}(x), h^{(l)}(x')) = e^{-\gamma_l \| a(h^{(l)}(x)) - a(h^{(l)}(x')) \|^2}\, .
\end{equation}

Given the random mini-batch $(X^b,Y^b)$ and the (randomly sub-sampled) half sub-batch $(X^c,Y^c)$, we evolve $\theta$ (and $\gamma$) in the steepest descent direction of the loss
{\small
\begin{equation}\label{KFlossWRN}
\begin{split}
   \L_\text{KF}(B) = \sum_{l=1}^4\lambda^{(l)} \|Y^b - K^{(l)}_{\gamma_l,\theta}(X^b,X^c) K^{(l)}_{\gamma_l,\theta}(X^c,X^c)^{-1} Y^c\|_2^2  + \L_\text{c-e}(f_\theta(X^b), Y^b)\, .
\end{split}
\end{equation}}

\subsubsection{Comparison to Dropout}

\begin{table}[h]
\begin{center}
\begin{tabular}{ | p{3.4cm} || p{2.0cm} | p{2.0cm} | p{2.0cm}| p{2.0cm}|}
\hline
\textbf{Training Method} & \textbf{Fashion MNIST} & \textbf{CIFAR-10} & \textbf{CIFAR-10.1} & \textbf{CIFAR-100}\\
\hline
BN & $4.95 \%$ & $4.68 \%$ & $10.85 \%$ & $20.39\%$\\

BN+KF & $4.90 \%$ & $4.53\%$ & $10.55 \%$ & $20.27\%$\\

BN+DO & $4.82 \%$ & $4.43\%$ & $10.50 \%$ & $19.50\%$\\

BN+DO+KF & $4.73\%$ & $4.22\%$ & $10.30\%$ & $19.20\%$\\
\hline

\end{tabular}
\end{center}
\caption{A comparison of the median test errors over 5 runs for networks trained on augmented data on Fashion MNIST, CIFAR-10, CIFAR-10.1, and CIFAR-100.  The second column to the right trains on augmented CIFAR-10 data but tests on CIFAR-10.1 data \cite{cifar101, 80M_tinyimages}.}\label{WRNKFvsDO}
\end{table}

Table \ref{WRNKFvsDO} compares the test errors obtained after training with only batch normalization (BN) with the incorporation of dropout (DO), KF, as well as a combination of all three.  The network architecture WRN-$16$-$8$ is used and median testing error over five runs is listed.  We train with step exponentially decreasing learning rates over $200$ epochs with identical hyperparameters as \cite{WRNZagKom}.  We observe that the addition of KF improves testing error rates against training with BN and BN+DO.  We also run a distributional shift experiment for CIFAR-10 using the data set CIFAR-10.1, \cite{cifar101} which is sampled from \cite{80M_tinyimages}.  As with the QMNIST experiment, we also observe improvements with the addition of KF.  %The trend line relating CIFAR-10.1 to the original CIFAR-10 testing error rates was established to have slope $1.62\pm 0.04$ across the various models and techniques.  In our WRN-$22$-$8$ trials, the ratio of the improvement of the use of KF training over standard BN training is $3.64$, while the corresponding ratio over DO training is $1.96$. Moreover, the corresponding ratios for WRN-$16$-$5$ are $1.83$ and $2$ respectively.

\begin{figure}[h]
	\begin{center}
			\includegraphics[width=\textwidth]{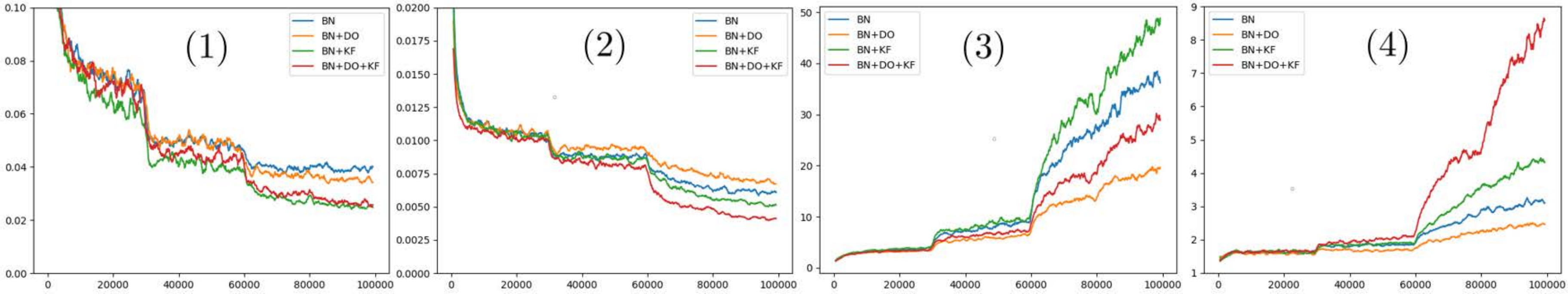}
		\caption{Single run using WRN-$16$-$8$ with each of BN only, BN+KF, BN+DO, and BN+KF+DO plotting (1) CIFAR-10 KF-loss (2) CIFAR-100 KF-loss (3) CIFAR-10 ratio of mean inter-class and in-class distances $h^{(4)}$ (4) CIFAR-100 ratio of mean inter-class and in-class distances $h^{(4)}$.}\label{WRNKF_lossratio}
	\end{center}
\end{figure}

We finally compare the KF loss, $\L_\text{KF}(B)$, and ratios of inter-class and in-class Euclidean distances on the output of the final convolutional layers within each batch in figure \ref{WRNKF_lossratio}.  These statistics are plotted over runs of WRN trained with CIFAR-10 and CIFAR-100.  We again observe reduced KF-losses and increased ratios of mean inter-class and in-class distances on the final convolutional layer output $h^{(4)}$ when comparing between BN and BN+KF as well as BN+DO and BN+DO+KF.  That is, KF reduces the distance (defined on the outputs of the inner layers) between images in the same class and increases that distance between images in distinct classes (thereby enhancing the separation).  The opposite effect is observed with the addition of dropout in training, suggesting they have differing reasons for improving testing error rates.

\begin{comment}
\section{Concluding remarks}

A framework for the application of kernel flows to Convolutional Neural Networks has been introduced which adds a generalizeability inspired loss function on hidden layer outputs.  This is optimized in tandem with the accuracy of the final output layer's accuracy.  We have presented a couple practical methods for its application to the MNIST database which has been shown to improve testing errors without decreasing training error as in dropout.  Further work could include research into richer kernel families possibly incorporating all hidden layer outputs.
\end{comment}

\section{Concluding remarks}
It has recently been found \cite{jacot2018neural, lee2019wide} that, in the overparameterized regime, training Neural Networks (or models) $f(x,\theta)$ with gradient descent and cross-entropy or mean squared loss is essentially equivalent to interpolating the data with the Neural Tangent Kernel $K(x,x')=\nabla_\theta f(x,\theta_0)\cdot \nabla_\theta f(x',\theta_0)$, i.e., when combined with gradient descent, losses defined as linear functionals (generalized moments) of the empirical distribution simply interpolate the data with a kernel fixed at initialization ($\theta_0$).
Kernel Flows on the other hand use non-linear functionals of the empirical distribution  designed to actually train the underlying kernel defined by the architecture of the Neural Network.
We suspect that these observations could to some degree explain the results observed in this paper (decreased generalization gap, improved test accuracies and increased robustness to distributional shift).

\paragraph{Acknowledgments}
The authors gratefully acknowledge support by  the Air Force Office of Scientific Research under award number FA9550-18-1-0271 (Games for Computation and Learning), Beyond Limits (Learning Optimal Models) and NASA/JPL (Earth 2050).

\bibliographystyle{plain}
\bibliography{extra}

\end{document}